# Minimizing Turns in Watchman Robot Navigation: Strategies and Solutions

Hamid Hoorfar
Department of Epidemiology and Public Health
University of Maryland
Baltimore, Maryland, USA
hhoorfar@som.umaryland.edu

Sara Moshtaghi Largani
Department of Computer Science
University of Cincinnati
Cincinnati, Ohio, USA
moshtasa@mail.uc.edu

Reza Rahimi
Department of Mathematics
University of Ottawa
Ottawa, Canada
rrahi080@uottawa.ca

Alireza Bagheri
Department of Computer Engineering
Amirkabir University of Technology
Tehran, Iran
ar.bagheri@aut.ac.ir

*Abstract*— **The Orthogonal Watchman Route Problem (OWRP) entails the search for the shortest path, known as the watchman route, that a robot must follow within a polygonal environment. The primary objective is to ensure that every point in the environment remains visible from at least one point on the route, allowing the robot to survey the entire area in a single, continuous sweep. This research places particular emphasis on reducing the number of turns in the route, as it is crucial for optimizing navigation in watchman routes within the field of robotics. The cost associated with changing direction is of significant importance, especially for specific types of robots. This paper introduces an efficient linear-time algorithm for solving the OWRP under the assumption that the environment is monotone. The findings of this study contribute to the progress of robotic systems by enabling the design of more streamlined patrol robots. These robots are capable of efficiently navigating complex environments while minimizing the number of turns. This advancement enhances their coverage and surveillance capabilities, making them highly effective in various real-world applications.**

*Keywords— path finding, robot navigation, robotics, algorithm*

## I. INTRODUCTION

In our rapidly evolving world, algorithms have assumed a paramount significance, continually expanding their role and influence. They have ingrained themselves into various aspects of our lives, shaping our daily routines and significantly impacting decision-making processes [1], system control [2], Artificial intelligence, sampling methods [3, 4], motion planning, graph theory [5, 6], and particularly robotics [7-9] and robot navigation stand as prime examples of how algorithms have become indispensable components of our modern existence. The current paper focuses specifically on the topic of robot navigation, delving into its intricacies and significance. Orthogonal watchman robot navigation plays a crucial role in various applications, including surveillance, monitoring, and security systems. The efficient traversal of the robot within a given environment is essential to ensure comprehensive coverage and effective monitoring. One key aspect of optimizing robot navigation is minimizing the number of turns or bends in the robot's path. Reducing turns not only improves the efficiency of movement but also minimizes the time required to complete a given task. In this paper, we address the challenge of minimizing turns in orthogonal watchman robot navigation. The goal is to develop effective strategies and propose practical solutions that can significantly reduce the number of bends in the robot's trajectory while maintaining complete visibility of the environment. By minimizing turns, we aim to enhance the overall efficiency, coverage, and surveillance capabilities of watchman robots. The Watchman Route Problem is an optimization problem in computational geometry that involves finding the shortest route for a watchman or robot to guard an entire area, given only the layout of the area. The problem is typically defined using a simple polygon to represent the area, and the objective is to find a shortest closed curve such that all points within the polygon and on its boundary are visible to at least one point on the curve. This problem can be approached in two different variants: the fixed variant, where the watchman passes through a specified boundary point, and the float variant, where there is no predetermined starting point for the watchman.

### A. Related works

Previous research has investigated various algorithms to address the Watchman Route Problem. Chin and Ntafos [10] developed an algorithm with a time complexity of $O(n^4)$ for the fixed variant, while Xuehou Tan [11] introduced an incremental solution with a time complexity of $O(n^3)$ that constructs the shortest route. For the float variant, Tan [12] and later Sagheer [13] contributed a solution with a worst-case time complexity of $O(n^5)$. Tan also proposed a linear-time 2-approximation algorithm for the fixed watchman route. In practical applications, watchmen are often implemented as robots, which have motion restrictions due to their structural limitations. Orthogonal robots, capable of movement in two perpendicular directions, offer cost-effective solutions and are commonly employed. Consequently, minimizing the number of bends in the route becomes crucial, as additional bends lead to increased costs. In the orthogonal variant of the Watchman Route Navigation, the area under consideration is an orthogonal polygon, and the shortest route must consist of an orthogonal polygonal curve with the fewest possible bends. Extensive research has been conducted in the field of guarding and securing orthogonal polygons [14-18], considering the prevalence of real-world scenarios that can be effectively represented using orthogonal polygonal environments [19-23].

These research efforts have significantly contributed to the development of more efficient solutions for such applications.

### B. Motivation

The motivation behind this research stems from the growing need for efficient navigation and surveillance in real-world scenarios that can be represented using orthogonal polygonal environments [24]. With the increasing prevalence of robotic systems and the desire to optimize their performance, it becomes crucial to address the challenges of minimizing turns and maximizing coverage in watchman robot navigation. By developing strategies and solutions specifically tailored to the Watchman Route Problem in orthogonal environments, this research aims to enhance the capabilities of watchman robots. The optimization of their routes, with a focus on minimizing the number of bends, can lead to cost-effective and time-efficient surveillance operations. This, in turn, enables improved coverage and security in various practical applications, ranging from building security to environmental monitoring. Furthermore, the outcomes of this research contribute to advancements in the field of computational geometry and robotic navigation. By exploring the unique characteristics of orthogonal polygons and leveraging the concept of fan polygons, novel algorithms can be developed to tackle the Watchman Route Problem. These algorithms offer valuable insights into the design and implementation of efficient navigation systems, allowing robots to navigate complex environments with reduced turns and improved overall performance. Overall, the motivation behind this research lies in the pursuit of optimizing watchman robot navigation in orthogonal polygonal environments, with the ultimate goal of enhancing their coverage, surveillance capabilities, and operational efficiency in real-world applications.

This paper proposes a linear-time algorithm to solve the float watchman route problem for monotone orthogonal polygons. Our algorithm achieves an exact linear-time solution and tackles the challenge of minimizing turns in orthogonal watchman robot navigation by presenting tailored strategies and solutions for the Watchman Route Problem. Our proposed linear-time algorithm provides efficient solutions for both the float and fixed watchman route problems within orthogonal monotone polygons. By reducing the number of bends, we aim to enhance the efficiency and coverage capabilities of watchman robots in real-world applications. The subsequent sections delve into the algorithm's details, experimental evaluations, and comparative analyses, providing valuable insights for researchers and practitioners in the field of robotic navigation.

## II. Preliminaries

During robot navigation in an environment, modeling the surroundings using geometric objects, such as polygons, proves advantageous. Polygons, particularly orthogonal ones, offer a simplified 2D representation of the environment, which greatly aids in the advancement of navigation algorithms, especially in urban settings. In this scenario, a robot can be viewed as a point or any other geometric object, thanks to the Minkowski summation, which eliminates the influence of dimensionality whether the objects are zero-dimensional or non-zero-dimensional. The environment of interest in this paper is an orthogonal x-monotone polygon denoted as P. A monotone polygon, also known as an x-monotone polygon, is a polygon where any line that is parallel to the y-axis intersects the polygon's boundary at most twice. This means that for any vertical line, it will intersect the polygon's edges at most twice. Monotone polygons have a special property that simplifies their analysis and makes them useful in various computational geometry algorithms. By extending the vertical edges incident to the reflex vertices of P, we can decompose it into rectangles. As a result, the decomposition of P leads to the formation of $(n-2)/2$ rectangles, where n represents the number of vertices. Let us denote this set of rectangles as $R = R_1, R_2, ..., R_m$, ordered from left to right based on the x-coordinate of their left edges when P is horizontal (refer to Fig. 1). The upper and lower horizontal edges of $R_i$ are denoted as $u_i$ and $l_i$, respectively. Additionally, $U = u_1, u_2, ..., u_m$ and $L = l_1, l_2, ..., l_m$ represent the lists of upper and lower horizontal edges, respectively, where $1 \le i \le m = (n-2)/2$. For a horizontal line segment s, $x(s)$ represents the x-coordinate of the left vertex of s, while $y(s)$ denotes the y-coordinate of the line segment. Notably, for all $1 \le i \le m - 1$, it holds true that $y(u_i) = y(u_{i+1})$ or $y(l_i) = y(l_{i+1})$. Building upon this observation, we define the edge of P that contains $u_i$ as $e(u_i)$ and the edge that contains $l_i$ as $e(l_i)$. To organize these edges from left to right, we introduce two sets: $EU = \{e(u_i) \mid 1 \le i \le m\}$ and $EL = \{e(l_i) \mid 1 \le i \le m\}$. Within the list of horizontal edges E of P, an edge $e_j$ is classified as a local maximum if $y(e_j) \ge y(e_{j-1})$ and $y(e_j) \ge y(e_{j+1})$. Similarly, an edge $e_k$ is considered a local minimum if $y(e_k) \le y(e_{k-1})$ and $y(e_k) \le y(e_{k+1})$. The edge $e(u_m)$ (or $e(l_m)$) is designated as a local maximum if $u_m$ (or $l_m$) represents a local maximum. Likewise, $u_n$ (or $l_n$) is termed a local minimum if $e(u_n)$ (or $e(l_n)$) is a local minimum. In the set R, $R_j$ is specifically identified as a local maximum if $u_j$ and $l_j$ represent a local maximum and a local minimum, respectively. We introduce the concept of weak visibility for two axis-parallel segments, l and k, they are considered weakly visible if an axis-parallel line segment can be drawn from a point on l to a point on k without intersecting boundary of P. Covering an ortho-convex polygon P can be accomplished quite easily by passing through a point located within its kernel. The kernel of the polygon refers to a region that allows for a comprehensive view of the entire shape. An ortho-convex polygon is defined as a polygon that maintains vertical and horizontal monotonicity, ensuring that its edges follow a consistent ascending or descending order along both axes.

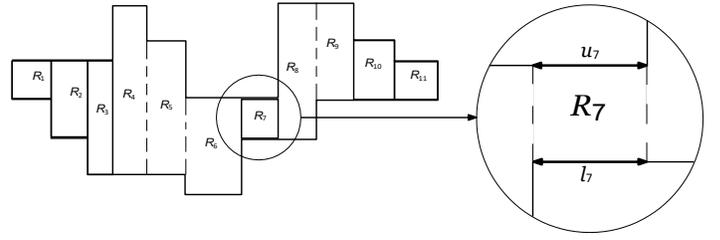

Fig. 1 - decompose an orthogonal to rectangles

## III. An Algorithm for Finding the Minimum Bends Orthogonal Route in a Monotone Environment

Consider an x-monotone orthogonal polygon, denoted as P. Our goal is to find an orthogonal route with the minimum number of bends while ensuring that every point within P is visible along the route. In other words, the route should allow for clear visibility of both the interior and boundary of P. A balanced polygon is an x-monotone orthogonal polygon that can be traversed with an orthogonal route represented by a horizontal line segment. It is characterized by a unique horizontal line segment known as the "align" segment, which connects the leftmost and rightmost vertical edges of the polygon without intersecting any other edges. It is important to note that there can be multiple align segments within a balanced polygon. Additionally, each balanced polygon contains an orthogonal corridor extending from its leftmost to rightmost edge, encompassing the align segment. Ortho-convex polygons are also classified as balanced polygons. Conversely, an orthogonal polygon that does not meet the criteria of a balanced polygon is referred to as an "unbalanced" polygon. Next, we present an algorithm for decomposing an x-monotone orthogonal polygon into balanced sub-polygons. The algorithm is based on identifying corridors within the polygon. Let ε represent the leftmost vertical edge of P. To find the first balanced part of P, start from ε and project a light beam perpendicular to ε and collinear with the X-axis. This rectilinear path of the light beam intersects a subset of rectangles, denoted as R, obtained from the vertical decomposition of P. We refer to this subset as $R_\rho$, which collectively forms a sub-polygon ρ and represents the first balanced part of the polygon. If $R_i$ is a rectangle within sub-polygon ρ, with $u_i$ and $l_i$ denoting the upper and lower horizontal edges of $R_i$, respectively, for every $u_i$ and $l_i$ belonging to ρ, it is established that:

$$min_{u_i \in \rho}(y(u_i)) \geq max_{l_j \in \rho}(y(l_j)) \qquad (1)$$

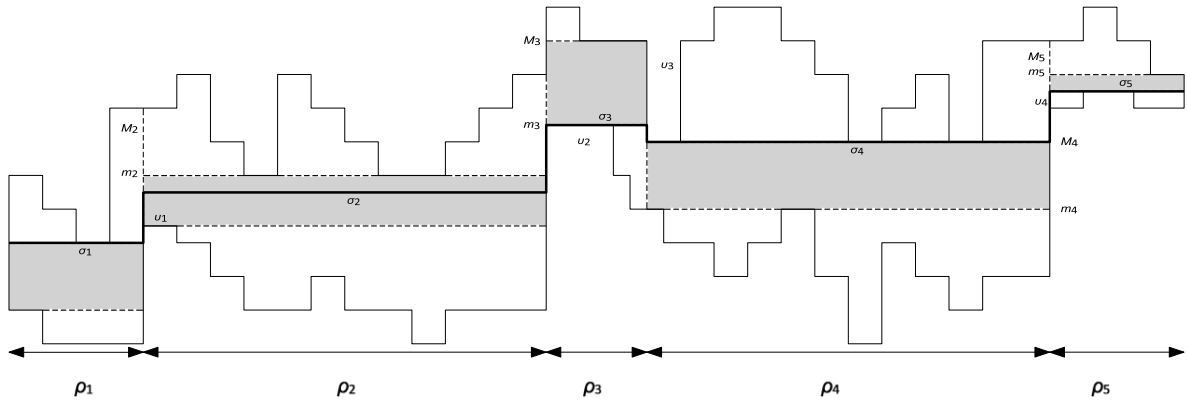

Fig. 2 - demonstration of selecting the suitable align segment for each balanced sub-polygon

This means that there exists a horizontal line segment σ that connects the leftmost and rightmost vertical edges of ρ, and σ is part of the balanced structure of ρ. By utilizing the align segment, we can determine an optimal orthogonal route for a balanced x-monotone polygon with the minimum length. To decompose P into multiple balanced x-monotone polygons, remove ρ from P and iterate through the aforementioned steps. This process yields a linear-time algorithm for decomposing P into balanced sub-polygons, as described in Algorithm 1.

When the condition in line 3 of Algorithm 1 is met, it signifies the existence of a balanced sub-polygon ρ. We proceed by removing ρ from P and iterating the algorithm for the remaining part of P, denoted as $P - \rho$. The rectangles belonging to ρ are removed from the set R, resulting in R being updated as $R = R - R\rho$. The elements in R are initially ordered from left to right and labeled accordingly. After removing σ, we relabel the remaining elements starting from 1. The same actions are

**Data:** *an x-monotone polygon with n vertices*

**Result:** *the shortest orthogonal route with minmum number of bends*

(1) *Set $min_u = u_l$ and $max_l = l_l$;*

(2) **foreach** *rectangle $R_i$ belongs to R* **do**

    (3) **if** *$u_i > max_l$ or $l_i < min_u$* **then**

        *remove $R_1, ..., R_{i-1}$ from R;*

        *refresh the index of R starting with 1;*

        *go to 1;*

    **end**

    (4) *Compute $min_u = min(min_u, u_i)$ and $max_l = max(max_l, l_i)$;*

**end**

Algorithm 1 - Decomposition P into the balanced sub-polygons

applied to the sets U and L. The number of iterations in this process is equal to the cardinality of R at the start. Hence, the time complexity for decomposing P into balanced polygons is linear. Each balanced polygon, such as ρ, possesses an align line-segment, denoted as σ, which connects its leftmost and rightmost edges. The align segment provides a view of the entire ρ from at least one point on σ, making ρ weak visible from σ. Therefore, σ becomes a potential route candidate for the orthogonal watchman route problem within ρ. Assuming P is decomposed into balanced sub-polygons $\rho_1, \rho_2, ..., \rho_k$, with align line-segments $\sigma_1, \sigma_2, ..., \sigma_k$ respectively, we can concatenate $\sigma_1, \sigma_2, ..., \sigma_k$ together using k - 1 vertical line-segments that connect the right end-point of $\sigma_i$ to the left end-point of $\sigma_{i+1}$, for every i between 1 and k-1. This resulting orthogonal path, denoted as Π, is the primary solution for the orthogonal floating watchman route problem in P. However, some excess portions can be trimmed from the beginning and end of Π to obtain the shortest possible orthogonal route. The watchman can ignore these trimmed parts and start guarding

**Data:** *an orthogonal path* Π

**Result:** *the shortest orthogonal route*

**(1) foreach** *rectangle $R_i$ belongs to R, from i=1 to m* **do**

> **(2)** **if** *$u_i$ is local maximum or $l_i$ is local minimum* **then**
>
> > *remove $R_1$, . . ., $R_i$ from R;*
> >
> > *go to item 3;*
>
> **end**

**end**

**(3) foreach** *or each rectangle $R_i$ belongs to R, from i=m down to 1* **do**

> **(4) if** *$u_i$ is local maximum or $l_i$ is local minimum*
>
> > *remove $R_i$, . . ., $R_m$ from R;*
> >
> > *go to item 5;*
>
> **end**

**end**

*(5) Compute* Π *=* Π *∩ R;*

Algorithm 2 - Trimming path Π to the shortest orthogonal route

from the leftmost point of the kernel of $\rho_1$ intersecting $\sigma_1$ to the rightmost point of the kernel of $\rho_k$ intersecting $\sigma_k$.

Now, we present a linear-time algorithm, described in Algorithm 2, for trimming Π to achieve the shortest orthogonal route. This algorithm focuses on removing unnecessary portions from Π while preserving its overall structure and properties.

**Data:** *a set of available aligns*

**Result:** *a set of appropriate aligns*

**foreach** *sub-polygon $\rho_1$* **do**

> **if** *$M_1 < M_2$* **then**
>
> > *set $y(\sigma_1) = M_1$;*
>
> **else**
>
> > *set $y(\sigma_1) = m_1$;*
>
> **end**

**end**

**foreach** *sub-polygon $\rho_i$ belongs to P, ordered from i = 1 to m* **do**

> **if** *$M_{i-1} < M_i$ and $M_i > M_{i+1}$* **then**
>
> > *set $y(\sigma_i) = m_i$;*
>
> **else**
>
> > *set $y(\sigma_i) = M_i$;*
>
> **end**

**end**

**foreach** *sub-polygon $\rho_m$* **do**

> **if** *$M_m < M_{m-1}$* **then**
>
> > *set $y(\sigma_m) = M_m$;*
>
> **else**
>
> > *set $y(\sigma_m) = m_m$;*
>
> **end**

**end**

Algorithm 3 - Align Selection

In most cases, this algorithm operates in constant time, but in the worst case, its complexity is O(n). However, the obtained route requires additional adjustments to become the minimum orthogonal route possible. These modifications focus on the lengths of the vertical line segments that connect consecutive align segments. Let $\Sigma = \{\sigma_1, \sigma_2, . . . , \sigma_k\}$ represent the align segments, and $y(\sigma_i)$ denote the y-coordinate of $\sigma_i$ for every i between 1 and k. Each sub-polygon $\rho_i$ has its align segment, and multiple horizontal line segments within the corridor of $\rho_i$ can serve as its align segment. Certain align segments are more favorable than others, resulting in a shorter route. The y-coordinate of an align segment can fall within two parameters of the balanced sub-polygon $\rho_i$, denoted as:

$$M_i = min_{u_j \in R_{\rho_i}} y(u_j) \tag{2}$$

and

$$m_i = max_{l_k \in R_{\rho_i}} \in y(l_k) \tag{3}$$

For every i between 1 and k, $M_i < m_{i+1}$ or $m_i > M_{i+1}$. When $M_{i-1} < M_i < M_{i+1}$ or $M_{i-1} > M_i > M_{i+1}$, the selection of whether $y(\sigma_i)$ equals $M_i$ or $m_i$ does not affect the total length of the vertical line segments connecting consecutive aligns $\sigma_{i-1}$, $\sigma_i$, and $\sigma_i$, $\sigma_{i+1}$ because it remains constant. However, when $M_{i-1} > M_i < M_{i+1}$ or $M_{i-1} > M_i < M_{i+1}$, the total length varies. In the former case, we choose the align segment with $y(\sigma_i) = m_i$, while in the latter case, we select the align segment with $y(\sigma_i) = M_i$. By doing so, the route becomes shorter as the total length of $\upsilon_{i-1}$ and $\upsilon_i$, where $\upsilon_{i-1}$ is the vertical line segment connecting the right endpoint of $\sigma_{i-1}$ and the left endpoint of $\sigma_i$, and $\upsilon_i$ is the vertical line segment connecting the right endpoint of $\sigma_i$ and the left endpoint of $\sigma_{i+1}$, becomes minimized. For the first balanced sub-polygon, if $M_1 < M_2$, we choose the align segment with $y(\sigma_1) = M_1$, and if $M_1 > M_2$, we select the align segment with $y(\sigma_1) = m_1$. Similarly, for the last sub-polygon $\rho_k$, if $M_k < M_{k-1}$, we choose the align segment with $y(\sigma_k) = M_k$, and if $M_k > M_{k-1}$, we select the align segment with $y(\sigma_k) = m_k$. Fig. 2 is for visualization.

We present a linear-time algorithm that selects the appropriate align segments for each balanced sub-polygon, aiming to achieve the shortest possible orthogonal route. The pseudo code of this algorithm is provided in Algorithm 3.

### A. Time Complexity

We have presented an algorithm that efficiently finds the shortest orthogonal path with the minimum number of bends. The algorithm is divided into three sections, each of which we have explained in detail. To obtain the desired results, it is crucial to run all three sections sequentially. Now, let's delve into the complexity of the algorithm. In the worst-case scenario, the time complexity of our algorithm is O(n). This means that the running time of the algorithm grows linearly with the number of edges in the polygon. We have achieved this efficient time complexity by designing the algorithm in a way that allows us to process each section in a linear manner. Therefore, the overall time complexity of the complete algorithm remains linear as well. This time complexity is optimal because, at the very least, to compute an orthogonal route, we must examine every vertex of the polygon. Hence, our algorithm is tightly optimized, and it is unlikely that a better time complexity can be achieved.

Additionally, the space complexity of the algorithm is O(n), meaning that the amount of memory required by the algorithm also grows linearly with the size of the input.

## IV. GENERALIZATION AND APPLICATION OF THE RESULT

We have extended our previous result to encompass orthogonal path polygons [15]. Path polygons are a type of orthogonal polygons where the dual graph induced by the vertical decomposition forms a path. Our exact linear-time algorithm for x-monotone orthogonal polygons can be utilized for vertical decomposition in this case as well. The vertical decomposition of an orthogonal polygon P involves dividing P into rectangles by extending the vertical edges of each reflex vertex inward until they intersect the boundary of P. The resulting dual graph represents the decomposition, with nodes corresponding to the rectangles and edges connecting adjacent rectangles [15]. Additionally, we define a fan polygon F as a simple polygon with a core vertex v, from which the entire polygon is visible. The core vertex v belongs to the kernel of the fan polygon, meaning that if a guard is stationed at v, the entire polygon is fully covered [15]. For an orthogonal polygon P with n vertices, if the dual graph of its decomposition forms a path, we classify it as a path polygon. Notably, all orthogonal monotones fall under the category of path polygons, and often the properties of path and monotone polygons align. Therefore, the algorithms we explained in the previous section can be utilized for path polygons as well. In other words, our algorithm can efficiently solve the problem in linear time for any simple orthogonal polygon P with a path dual graph. While the technique described in [15] involves converting a path polygon into a monotone through "unfolding," we present an alternative strategy for finding the shortest orthogonal route in path polygons [see Fig. 3]. The distinction between path polygons and monotone polygons lies in the presence of reflex rectangles. Reflex rectangles are rectangle components of the vertical decomposition where both neighboring components are located on the same side, either the left or the right. Other rectangle components have neighbors on different sides, with one on the left and another on the right, except for rectangles R1 and Rm, which have only one neighbor. Initially, we remove all reflex rectangles from P, leading to the decomposition of the path polygon into several monotone sub-polygons. We then apply our algorithm, consisting of decomposing each obtained monotone sub-polygon into balanced polygons and selecting appropriate aligns. Connecting the resulting routes from the balanced sub-polygons is straightforward, requiring the drawing of vertical line segments between their endpoints. After trimming, the obtained route represents the shortest possible route.

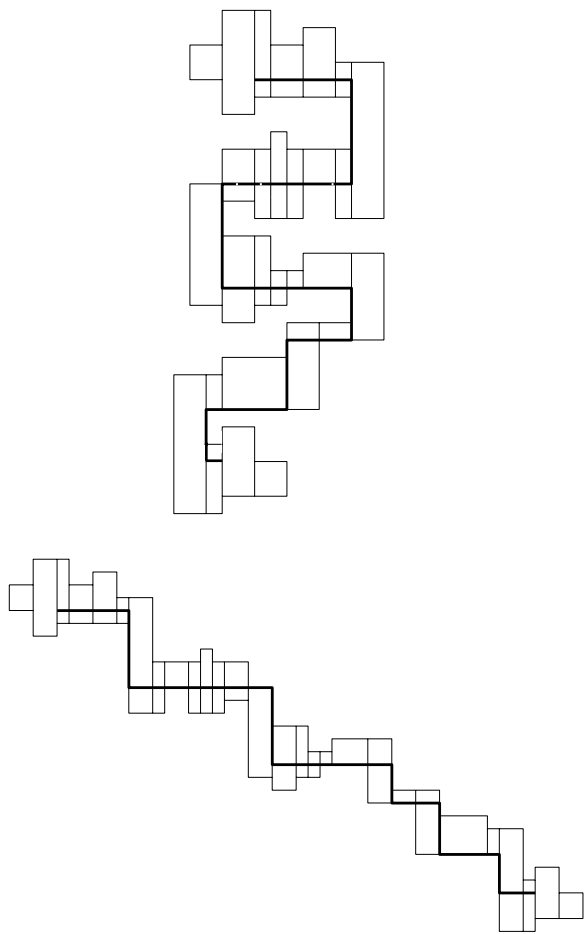

## V. CONCLUSION

In conclusion, this research has addressed the OWRP by proposing a linear-time algorithm for finding a minimum-length watchman route within a monotone polygonal environment. The focus of the algorithm is to minimize the number of turns in the route, which is crucial for optimizing watchman robot navigation. By providing a solution to the OWRP, this research contributes to the advancement of robotic systems, particularly in the field of watchman robots. These robots play a vital role in various applications, such as surveillance and coverage in complex environments. The proposed algorithm enables the design of more efficient watchman robots capable of navigating through intricate environments with minimal turns. The results of this research have significant implications for real-world applications, where effective coverage and surveillance are crucial. By minimizing the number of turns in the watchman route, the proposed algorithm improves the efficiency and effectiveness of watchman robots in fulfilling their tasks.

Further research can be conducted to explore extensions and adaptations of the algorithm for non-monotone environments. Additionally, investigating the integration of other optimization criteria, such as energy efficiency or obstacle avoidance, could enhance the algorithm's applicability in diverse scenarios. Overall, the presented linear-time algorithm for the OWRP provides a valuable contribution to the field of robotics, offering a practical solution for optimizing watchman robot navigation and enabling more effective and efficient surveillance and coverage in complex environments.

Fig. 3 - Unfolding a path polygon and obtaining a monotone polygon.